\documentclass[graybox]{svmult}

\usepackage{amsmath}
\usepackage{amssymb}
\usepackage{algorithm, algpseudocode}
\usepackage{type1cm}        
\usepackage{accents}
\newcommand{\ubar}[1]{\underaccent{\bar}{#1}}

\usepackage[framemethod=tikz]{mdframed}
\usepackage{makeidx}         
\usepackage{graphicx}        
\usepackage{multicol}        
\usepackage[bottom]{footmisc}

\usepackage[numbers]{natbib}
\usepackage{newtxtext}       %
\usepackage{newtxmath}       

\usepackage{tikz}
\usetikzlibrary{backgrounds}
\usepackage{pgfplots}        
\usepgfplotslibrary{dateplot}

\newmdenv[
    backgroundcolor=gray!15,
    linecolor=gray!15,
    innertopmargin=3pt,
    innerbottommargin=3pt,
    skipabove=0pt,
    skipbelow=0pt
]{examplebox}

\newcommand{\R}{\mathbb{R}}
\newcommand{\EE}{\mathbb{E}}
\newcommand{\Z}{\mathbb{Z}}

\makeindex             

\pgfplotsset{compat=1.16}
\begin{document}
\setcitestyle{square}

\title*{Dynamic Gaussian Processes and the Vanilla-SPDE Exchange}
\titlerunning{Dynamic Gaussian Processes and the Vanilla-SPDE Exchange}
\author{Rui-Yang Zhang, Lachlan Astfalck, Edward Cripps, David Leslie, Henry Moss}
\authorrunning{Zhang, Astfalck, Cripps, Leslie, Moss}
\institute{Rui-Yang Zhang \at School of Mathematical Sciences, Lancaster University. \\
e-mail: r.zhang26@lancaster.ac.uk 
\and Lachlan Astfalck \at School of Mathematics and Statistics, The University of New South Wales.\\
e-mail: l.astfalck@unsw.edu.au
\and Edward Cripps \at School of Physics, Mathematics and Computing, The University of Western Australia.\\
e-mail: edward.cripps@uwa.edu.au
\and David Leslie \at School of Mathematical Sciences, Lancaster University.\\ 
e-mail: d.leslie@lancaster.ac.uk 
\and Henry Moss \at School of Mathematical Sciences, Lancaster University.\\
e-mail: henry.moss@lancaster.ac.uk 
}

%
%
\maketitle

\abstract{Gaussian process inference is often limited by cubic computational costs, a challenge that becomes more pronounced in spatio-temporal settings where posterior inference is required over dense grids. While state-space SPDE formulations enable linear complexity in time, exact inference remains cubic in space and deteriorates further when observation locations are disjoint from the prediction locations, which inflates the number of considered spatial points. To address this, we propose the \textit{Vanilla-SPDE Exchange}, which exploits an equivalence between the standard and SPDE formulations of GP inference to construct a hybrid scheme with improved computational cost. We demonstrate these gains through complexity analysis and numerical experiments.}


\section{Introduction}

Gaussian processes (GPs) are stochastic process models widely used for describing spatio-temporal phenomena \citep{wikle2019spatio} due to their expressiveness through flexible covariance function choices and abilities to quantify uncertainties naturally. 

The standard GP inferences jointly consider all the observation and prediction points for computations such as likelihood evaluations, obtaining posterior distributions, and posterior sampling \citep{williams2006gaussian}. The joint consideration of all relevant points implies a \textit{static} view of the stochastic process, and induces the cubic scaling of computational complexity in the number of considered points. 

On the other hand, Stochastic Partial Differential Equation (SPDE) formulations \citep{sarkka2013spatiotemporal} of certain classes of spatio-temporal Gaussian processes take a \textit{dynamic} approach to the stochastic process by treating the GP as the stationary solution of a SPDE. Furthermore, these SPDEs admit closed-form transition densities in time, so one can conduct the exact inference sequentially. Thus, the implication of the dynamic formulation is that the computations that scale cubically in complexity under the static formulation, while still scaling cubically in the number of spatial points, only scale linearly in the number of temporal points. 

However, in many spatio-temporal inference problems, the observation and prediction locations are spatially disjoint. For example, predictions may be required at a fixed set of locations over time, while observations are collected elsewhere. In such settings, the SPDE formulation must include both the observation and prediction locations in the state space. Consequently, the number of spatial locations can increase substantially, and the resulting cubic dependence on the spatial dimension may outweigh the linear-in-time advantage of sequential inference. As a result, SPDE-based inference can become more computationally expensive than standard GP inference.

In this work, we propose a new GP inference technique, called the \textit{Vanilla-SPDE Exchange} (VaSE), that allows the full GP computation to be decomposed into components, each tackled by either the static or dynamic formulation, to obtain the best of both worlds. In particular, for the computation of posterior sampling of spatio-temporal GPs with disjoint observation-prediction locations, the hybrid inference of VaSE regresses the observations with the static method, then predicts and propagates the posterior samples using the dynamic method. 

This paper is organised as follows. The static Gaussian process inference is described in Section \ref{sec:static}. The dynamic Gaussian process framework and its sequential inference are then explored in Section \ref{sec:dynamic}, which motivates the vanilla-SPDE exchange in Section \ref{sec:vase}. Finally, the benefits of posterior sampling using VaSE for settings with disjoint observation-prediction locations are numerically investigated in Section \ref{sec:experiments}.

\section{Static Gaussian Process Inference} \label{sec:static}

A \textit{Gaussian process} (GP) $\{f(x)\}_{x\in X}$ is a stochastic process in which every linear combination $\sum_{j}a_jf({x_j})$, for real coefficients $a_j \in \R$ and input points $x_j \in X$, is a multivariate Gaussian distribution \citep{williams2006gaussian}. It is standard to characterise a Gaussian process by its \textit{mean} and \textit{covariance} functions, i.e. $f \sim \mathrm{GP}(\mu, k)$ where $\EE[f(x)] = \mu(x)$ and $\mathrm{Cov}(f(x), f(x')) = k(x, x')$ for any $x, x' \in X$. The Gaussianity requirement of a GP thus demands the covariance function (also called a \textit{kernel}) $k$ to be a \textit{positive semi-definite} function \citep{Lindgren2012}. 

In this paper, we assume the GP is \textit{spatio-temporal}, so the domain of the process is the Cartesian product $X = S \times T$ of spatial $S$ and temporal $T$ components. We use $\boldsymbol{x} = (s,t) \in X$ to denote a point with location $s$ and time $t$. Given data $\mathcal{D} = \{(\boldsymbol{x}_i, y_i)\}_{i = 1}^N = \{ \ubar{\boldsymbol{x}}, \ubar{y}\}$ as noisy observations of the truth, i.e. $y_i = f(\boldsymbol{x}_i) + \varepsilon_i$, $\varepsilon_i \sim N(0, \sigma_{\text{obs}}^2)$, the posterior distribution of $f$ at any vector of $M$ test points $\ubar{\boldsymbol{x}}^* \in X^M$ is given by \begin{equation} \label{eqn:static-gp-regression}
\begin{split}
    f(\ubar{\boldsymbol{x}}^*) | \mathcal{D} &\sim N(\mu_{\ubar{y}^*|\mathcal{D}}, \Sigma_{\ubar{y}^*|\mathcal{D}}), \\ 
    \mu_{\ubar{y}^*|\mathcal{D}}&= \mu(\ubar{\boldsymbol{x}}^*) + K_*^T (K+\sigma_{\text{obs}}I_N)^{-1}(\ubar{y} - \mu(\ubar{y})) \\
    \Sigma_{\ubar{y}^*|\mathcal{D}}&= K_{**} - K_*^T (K+\sigma_{\text{obs}}I_N)^{-1} K_*
\end{split}
\end{equation}
with $N\times N$ identity matrix $I_N$ and Gram matrices $K = (k(\boldsymbol{x}_i, \boldsymbol{x}_j))_{ij} \in \mathbb{R}^{N \times N}$, $K_* =(k(\boldsymbol{x}_i, \boldsymbol{x}^*_j))_{ij} \in \R^{N \times M}$, $K_{**} =  (k(\boldsymbol{x}^*_i, \boldsymbol{x}^*_j))_{ij} \in \R^{M\times M}$. Furthermore, if one wishes to draw a sample from this posterior $\ubar{y}^* \sim f(\ubar{\boldsymbol{x}}^*) | \mathcal{D}$, it can be achieved by scaling and shifting an $M$-dimensional standard Gaussian vector, i.e. \begin{equation}\label{eqn:static-gp-sample}
\ubar{y}^* = \mu_{\ubar{y}^*|\mathcal{D}} + \sqrt{\Sigma_{\ubar{y}^*|\mathcal{D}}} \xi,  \qquad \xi \sim N(0, I_M)
\end{equation}
where $\sqrt{\Sigma_{\ubar{y}^*|\mathcal{D}} }$ is the matrix square root.

Notice that the Gram matrices used in \eqref{eqn:static-gp-regression} represent the joint dependency between observation points and prediction points. Rather than updating predictions recursively as new data arrive, the GP constructs a covariance matrix considering every training and prediction location at once, and predictions are obtained through a marginal conditioning operation. So, test points are not treated as future states evolving through time, but simply as additional coordinates in a large correlated random vector. This thus indicates the static nature of the standard GP formulation.

Formulas such as \eqref{eqn:static-gp-regression} and \eqref{eqn:static-gp-sample} indicate the cubic complexity scaling of GP regressions and sampling. Following standard matrix computation complexity analysis \citep{golub2013matrix}, for matrices $A \in \R^{N \times N}$ and $B \in \R^{N \times M}$, computations such as matrix inversion $A^{-1}$ is of complexity $O(N^3)$, matrix square root (or Cholesky decomposition) $\sqrt{A}$ is of complexity $O(N^3)$, and matrix multiplication $A B$ is of complexity $O(N^2M)$. Therefore, the computation of the posterior covariance matrix, following \eqref{eqn:static-gp-regression}, is of complexity $O(N^3 + N^2M + N M^2)$, while drawing $J$ samples from the posterior following \eqref{eqn:static-gp-sample} costs an additional $O(J M^3)$ after obtaining the posterior distribution's mean vector and covariance matrix. 

In addition, in the spatio-temporal setting, we would often be interested in predicting on the same set of spatial locations over time, so the $M$ test points consist of $M_s$ spatial locations and $M_t$ time points such that $M = M_s M_t$. Therefore, the full computational complexity to draw $J$ samples from the posterior, including regression, is $O(N^3 + N^2M_s M_t + N M_s^2 M_t^2 + JM_s^3 M_t^3)$, which could be formidable when any of the observation number $N$, prediction spatial location $M_s$, and prediction time points $M_t$ is large. 

A common remedy for the costly GP computations is to conduct approximate inferences by reducing the effective number of considered observation locations. Such approximate methods include inducing points \citep{titsias2009variational} and Vecchia approximation \citep{katzfuss2021general}. In this paper, however, we focus on exact GP inference, although the methods we propose here are amenable to further approximation and computational speed-up, as we will explain further in Section \ref{sec:vase}.  

\section{The Dynamic approach to Gaussian Process Inference} \label{sec:dynamic}

An alternative exact GP inference method with improved scalability is the dynamic approach \citep{hartikainen2010kalman, sarkka2012infinite}. For a wide class of Gaussian processes, the stochastic process can be formulated as the stationary solution to a stochastic (partial) differential equation (SPDE) with closed-form transition densities across time so as to admit sequential inference that scales linearly in time. 

The class of such processes includes spatio-temporal GPs with Mat\'ern temporal kernel that is separable from the spatial component. In this section, we will explore how such processes can be described as the stationary solution to an SPDE. We begin by introducing how the time domain GP, such as a one-dimensional Mat\'ern, can be viewed as the stationary solution to a stochastic differential equation (SDE) \citep{oksendal2003stochastic}, which enables sequential inference in Section \ref{sec:dynamic-gp-temporal}. Subsequently, we extend the argument to spatio-temporal GPs with separable kernels in Section \ref{sec:dynamic-gp-spatio-temporal}. A computational complexity comparison between the static and dynamic formulations is then presented in Section \ref{sec:dynamic-gp-complexity} to motivate why further improvements are required for some spatio-temporal computations. 

\subsection{Time Domain GP}\label{sec:dynamic-gp-temporal}

We start by considering a one-dimensional Gaussian process $\{f(t)\}_{t \in \R}$ with zero-mean and covariance $k$, i.e. $f \sim \mathrm{GP}(0, k)$. We further assume this GP is \textit{stationary} so that the covariance between two points is only dependent on the distance between the two points, i.e. $k(t, t') = k(|t-t'|)$. A recurring example of such a one-dimensional stationary kernel is that of the Mat\'ern kernel with half-integer (i.e. $p + 1/2$ for $p \in \Z$) smoothness \citep{guttorp2006studies, porcu2024matern}. In the special cases of $p = 1, 2$, we have the following expressions of Mat\'ern 3/2 and 5/2 kernels:\[
\begin{split}
k_{\text{Mat}-3/2}(t,t') &= \left( 1 + \frac{\sqrt{3} |t-t'|}{l}\right) \exp \left[ - \frac{\sqrt{3} |t-t'|}{l} \right], \\
k_{\text{Mat}-5/2}(t,t') &= \left( 1 + \frac{\sqrt{5} |t-t'|}{l} + \frac{5 |t-t'|^2}{3l^2}\right) \exp \left[ - \frac{\sqrt{5} |t-t'|}{l} \right]. 
\end{split}
\]
A Mat\'ern $(p + 1/2)$ Gaussian process $\{f(t)\}$ is the stationary solution to the SDE \[
\left( \kappa + \frac{d}{dt}\right)^{p+1} f(t) = W(t)
\]
with a white noise process of a suitable variance \citep{hartikainen2010kalman, doob1944elementary,rey2006open, whittle1954stationary}. In the special case of $p = 1$, the Mat\'ern-$3/2$ GP of lengthscale $l$ and variance $\sigma$ is the stationary solution of the SDE \begin{equation}\label{eqn:matern-3/2-sde}
\begin{bmatrix}
df(t) \\
df'(t)
\end{bmatrix} = \underbrace{\begin{bmatrix}
0 & 1 \\
- \kappa^2 & -2\kappa
\end{bmatrix}}_F \begin{bmatrix}
f(t) \\
f'(t)
\end{bmatrix} dt + \kappa^{3/2} \sigma dW(t)
\end{equation}
where $\kappa = \sqrt{3}/l$. This linear SDE admits an exact transition when initialised at stationarity: with $\boldsymbol{f} = [f, f']^T$, we have \[
\begin{split}
    \boldsymbol{f} ( t+ \delta) &= \Phi  \boldsymbol{f}(t) + \eta_t \\
    \Phi  = \exp[ F\delta], \quad \eta_t \sim N_2(0, Q), \quad &Q = P_\infty - \Phi P_\infty \Phi^T, \quad P_\infty = \mathrm{diag}[\sigma^2, \kappa^2 \sigma^2].
\end{split}
\]
which follows from the stationarity of the process. Note that the transition matrices $\Phi, Q, P_\infty$ depend on the stepsize $\delta$, and we omit this dependency in the notation for simplicity in the following expositions. 

We wish to conduct regression using noisy observations $\{y(t_k)\}_{k = 1}^N$ where $y(t_k) = f(t_k) + \varepsilon_k$ and $\varepsilon_k \sim N(0, \sigma_{\mathrm{obs}}^2)$ for all $k = 1, 2, \ldots, N$. Due to the SDE formulation of the considered Gaussian process, we construct the regression as a \textit{state space model} \citep{brockwell2016introduction} where the SDE is the latent process with observations $\{y(t_k)\}_{k = 1}^N$ at locations $\{t_k\}_k$. In particular, we have the following linear Gaussian state space model:
\[
\begin{aligned}
\boldsymbol{f}_{k+1} &= \Phi \boldsymbol{f}_k + \eta_k, \qquad \eta_k \sim N_2(0, Q), \\
y_k &= G \boldsymbol{f}_k + \varepsilon_k, \qquad \varepsilon_k \sim N(0, \sigma_{\mathrm{obs}}^2),
\end{aligned}
\]
where $y_k := y(t_k)$, $\boldsymbol{f}_k := \boldsymbol{f}(t_k)$, and $G = \begin{bmatrix} 1 & 0 \end{bmatrix}$. 

A key property of the linear Gaussian model is that its predicting, filtering and smoothing distributions are all Gaussians, i.e. we have for any $k = 1, 2, \ldots, N$, 
\begin{itemize}
    \item (predicting distribution) $\boldsymbol{f}_{k+1} | y_{1:k} \sim N(\boldsymbol{m}_{k+1|k}, \boldsymbol{P}_{k+1|k})$
    \item (filtering distribution) $\boldsymbol{f}_k | y_{1:k} \sim N(\boldsymbol{m}_{k|k}, \boldsymbol{P}_{k|k})$
    \item (smoothing distribution) $\boldsymbol{f}_k | y_{1:N} \sim N(\boldsymbol{m}_{k|N}, \boldsymbol{P}_{k|N})$
\end{itemize}
and these distributions' mean vectors and covariance matrices can be obtained recursively. We also note that the smoothing distributions correspond to the posterior distributions at index $k$. 

The mean $\boldsymbol{m}_{k|k}$ and covariance $\boldsymbol{P}_{k|k}$ of the filtering distribution $\boldsymbol{f}_k | y_{1:k}$ can be calculated using the Kalman filter. For each $k = 1, 2, \ldots, N$, we have the following standard formula \citep{brockwell2016introduction}.  
\[
\begin{aligned}
\textbf{Prediction:}\quad 
&\boldsymbol{m}_{k+1|k} = \Phi \boldsymbol{m}_{k|k}, \qquad
\boldsymbol{P}_{k+1|k} = \Phi \boldsymbol{P}_{k|k} \Phi^T + Q, \\[4pt]
\textbf{Update:}\quad 
&v_{k+1} = y_{k+1} - G \boldsymbol{m}_{k+1|k}, \qquad
S_{k+1} = G \boldsymbol{P}_{k+1|k} G^T + \sigma_{\mathrm{obs}}^2, \\
&K_{k+1} = \boldsymbol{P}_{k+1|k} G^T S_{k+1}^{-1}, \qquad
\boldsymbol{m}_{k+1|k+1} = \boldsymbol{m}_{k+1|k} + K_{k+1} v_{k+1}, \\
&\boldsymbol{P}_{k+1|k+1} = \boldsymbol{P}_{k+1|k} - K_{k+1} G \boldsymbol{P}_{k+1|k}.
\end{aligned}
\]

After filtering through all $N$ observations $y_1, \ldots, y_N$, we can obtain the mean $\boldsymbol{m}_{k|N}$ and covariance $\boldsymbol{P}_{k|N}$ of the smoothing distribution $\boldsymbol{f}_k | y_{1:N}$ using the Rauch--Tung--Striebel (RTS) smoother \citep{brockwell2016introduction}. For $k = N-1, N-2, \ldots, 0$, we have the following standard recursion. \[
\begin{aligned}
\textbf{RTS:}\quad 
&J_k = \boldsymbol{P}_{k|k} \Phi^T \left(\boldsymbol{P}_{k+1|k}\right)^{-1}, \\
&\boldsymbol{m}_{k|N} = \boldsymbol{m}_{k|k} + J_k \left(\boldsymbol{m}_{k+1|N} - \boldsymbol{m}_{k+1|k}\right), \\
&\boldsymbol{P}_{k|N} = \boldsymbol{P}_{k|k} + J_k \left(\boldsymbol{P}_{k+1|N} - \boldsymbol{P}_{k+1|k}\right) J_k^T.
\end{aligned}
\]
From the above formulas, we can notice that GP computations now scale linearly in the number of observation times. Likelihood evaluation can be achieved by doing the $N$ steps of the Kalman filter, thus it is $O(N)$. Posterior regression and sampling at $M$ prediction points can similarly be achieved in $O(N+M)$ complexity to complete both the filtering and smoothing steps. This is in stark contrast to the cubic-scaling complexities of the static formulation we saw in Section \ref{sec:static}. 

Note that we presented the sequential inference formulas for regression when observations are made with a regular time increment $\delta_t$. Non-regular time increments are possible, although one then needs to adjust the matrices $\Phi$ and $Q$ accordingly. 

\subsection{Extending to Spatio-Temporal Settings}\label{sec:dynamic-gp-spatio-temporal}

The sequential inference framework above extends naturally to Gaussian processes with separable kernels. Here, we consider a spatio-temporal GP with a kernel
\[
k_\text{full}((s,t),(s',t')) = k_{s}(s, s') k_{t}(t, t'),
\]
where $k_{t}$ is a Mat\'ern half-integer kernel and $k_{s}$ is arbitrary. Following \citep{sarkka2013spatiotemporal}, we can formulate such a GP as the stationary solution to a Stochastic Partial Differential Equation (SPDE). For Mat\'ern-3/2 temporal and any spatial kernel $k_s$, we have the SPDE \[
d\boldsymbol{f}(t) = (I_{N_s} \otimes F)\boldsymbol{f}(t)\, dt + (I_{N_s} \otimes L)\, d\boldsymbol{W}(t).
\]
where $L = [0, 1]$, $N_s$ is the total number of spatial locations considered, and $\otimes$ is the Kronecker product. This is closely related to the temporal SDE of \eqref{eqn:matern-3/2-sde} with states extended due to the additional spatial components. Furthermore, with $K_s \in \R^{N_s \times N_s}$ representing the Gram matrix of all considered locations,  the corresponding state space model of the above SPDE is
\[
\begin{split}
\boldsymbol{f}_{k+1} &= (I_{N_x} \otimes \Phi)\boldsymbol{f}_k + \eta_k, 
\qquad \eta_k \sim N(0,\, K_{x} \otimes Q),\\
y_k &= G_k \boldsymbol{f}_k + \varepsilon_k, 
\qquad \varepsilon_k \sim N(0, \sigma_{\mathrm{obs}}^2 I_k),
\end{split}
\]
where $G_k$ is the selection matrix for the states observed at time index $k$. This yields a linear Gaussian model of identical form to the temporal case, with the substitutions
\[
\Phi \;\mapsto\; I_{N_s} \otimes \Phi, 
\qquad Q \;\mapsto\; K_{s} \otimes Q, 
\qquad G \;\mapsto\; G_k.
\]
Sequential inference, therefore, proceeds exactly as before using Kalman filtering and RTS smoothing. The inclusion of the extended $N_s$ spatial components inflates the computational costs of each sequential inference step, making the computational complexity scale linearly in time but cubically in space. For example, likelihood evaluation costs $O(N_s^3 N_t)$, which is still a noticeable improvement from the static formulation's $O(N_s^3N_t^3)$ cost. However, we will see below that the comparison is less clear-cut for regression and posterior sampling. 

\subsection{Complexity Comparison of Static and Dynamic GPs} \label{sec:dynamic-gp-complexity}

Although the dynamic GP and its sequential inference promise computational speed-ups, it is not always beneficial. Here, we compare the computational costs under static kernel-based GP and dynamic SPDE-GP for different computations under a general spatio-temporal inference set-up to illustrate the issue. 

Consider conducting inference with a spatio-temporal Gaussian process $\{f(s, t)\}$ with a separable kernel such that the temporal component admits the dynamic formulation. Denote the data points as $\mathcal{D} = \{ s_i, t_i, y_i \}_{i = 1}^N$ with $N_s$ unique spatial locations and $N_t$ unique temporal locations; the test points as $\mathcal{S} = \{ s^*_j, t^*_j \}_{j = 1}^M$ with $M_s$ unique spatial locations and $M_t$ unique temporal locations; and the data-test combined points $\{ x_u, h_u \}_{u = 1}^U$ with $U_s$ unique spatial locations and $U_t$ unique temporal locations. By construction, we have $U \ge N, M$, $U_s \ge N_s, M_s$, and $U_t \ge N_t, M_t$ respectively. Also, if the data are additionally made on a fixed spatial grid, then we have $N = N_s N_t$. Similarly, $M = M_s M_t$ when test points are on a spatio-temporal grid.

Under this set-up, the SPDE-GP formulation as described in Section \ref{sec:dynamic-gp-spatio-temporal} will have to consider all $U_s$ spatial points at once in the sequential inference of $U_t$ steps. This induces the $O(U_s^3 U_t)$ complexity. For posterior sampling at those points, a further $O(M_s^2M_t)$ cost is needed for each sample to be propagated through $M_t$ steps by multiplying $M_s \times M_s$ matrices with a $M_s$-dimensional state vector. 

\renewcommand{\arraystretch}{1.2}
\begin{table}[ht!]
    \centering
    \begin{tabular}{|c|c|c|}
    \hline
    \textbf{GP Computations} & \textbf{Kernel-Based GP Cost } $O(\cdot)$ & \textbf{SPDE-GP Cost } $O(\cdot)$ \\
    \hline
    Likelihood Evaluation with Data $\mathcal{D}$
    & $N^3$ 
    & $N_s^3 N_t$ \\
    \hline
    Posterior Distribution on Test Points $\mathcal{S}$
    & $N^3 + N^2 M + N M^2$ 
    & $U_s^3 U_t$ \\
    \hline
    $J$ Posterior Samples on Test Points $\mathcal{S}$
    & $N^3 + N^2 M + N M^2 + J M^2$ 
    & $U_s^3 U_t + J M_s^2 M_t$ \\
    \hline
    \end{tabular}
    \caption{Computational complexity comparison between kernel-based GP and SPDE-based GP.}
    \label{tab:vase-computational costs}
\end{table}

The computational costs are summarised in Table \ref{tab:vase-computational costs}. Importantly, neither formulation uniformly dominates the other. Their relative efficiency depends on the spatial and temporal structure of the observations and prediction locations. The following examples demonstrate how different scenarios lead to varying relative performances of the methods. 

\begin{enumerate}

\item \textbf{Fixed observation locations over time.}

Suppose observations are collected repeatedly at the same set of spatial locations, so that $N = N_s N_t$ with $N_s \ll N$. In this case, the SPDE-GP likelihood evaluation cost $O(N_s^3 N_t)$ can be substantially smaller than the kernel-based GP cost $O(N^3) = O(N_s^3 N_t^3)$. If prediction locations largely overlap with the observation locations, then $U_s \approx \max\{N_s, M_s\}$, and the SPDE-GP posterior computation $O(U_s^3 U_t)$ also benefits from its linear dependence on the temporal dimension. This is the setting in which the dynamic formulation is most advantageous.

\item \textbf{Moving observation locations over time.}

Suppose observations are collected at different spatial locations at each time point. Then the number of unique spatial locations may grow with the sample size, giving $N_s \approx N$, unlike the gridded setup above. The SPDE-GP likelihood cost becomes $O(N_s^3 N_t) \approx O(N^3 N_t)$, while the kernel-based GP remains $O(N^3)$. In this regime, the dynamic formulation loses its computational advantage because sequential inference must propagate an increasingly large spatial state vector.

\item \textbf{Large prediction grids and posterior sampling.}

Consider a situation where posterior samples are required on a dense spatio-temporal prediction grid, so that $M = M_s M_t$ is large. Generating samples from a kernel-based GP requires $O(M^2)$ work per sample, whereas SPDE-GP sampling requires only $O(M_s^2 M_t)$ by propagating the state sequentially through time. Since $M^2 = M_s^2 M_t^2 \gg M_s^2 M_t$ whenever $M_t$ is large, SPDE-GP is often substantially more efficient for posterior sample generation, even when its posterior computation itself is not cheaper.

\item \textbf{Disjoint observation and prediction locations.}

When the observation locations and prediction locations have little or no spatial overlap, $U_s$ approaches the total number of unique locations across both sets. Consequently, the SPDE-GP posterior computation $O(U_s^3 U_t)$ can become expensive, while the kernel-based GP still incurs the large dense-matrix costs associated with $N$ and $M$. In such cases, neither method is uniformly efficient across all stages of inference. An example of such a setup can be found in \citep{zhang2025ballast}.

\end{enumerate}

These examples suggest that the two formulations excel at different tasks. Kernel-based GPs are often preferable for regression when observations are collected at many distinct spatial locations, whereas SPDE-GPs are particularly attractive for forecasting and posterior sample generation due to their linear scaling in the temporal dimension. This motivates the Vanilla-SPDE Exchange (VaSE), which combines the strengths of both approaches while incurring minimal additional computational cost, as described in Section \ref{sec:vase}.

\section{The Vanilla-SPDE Exchange} \label{sec:vase}

The preceding complexity analysis suggests that kernel-based and SPDE-based GP inference possess complementary computational advantages. In particular, kernel-based GPs are often preferable for regression when observations are collected at many distinct spatial locations, whereas SPDE-GPs are particularly attractive for forecasting and posterior sample generation due to their linear scaling in the temporal dimension. This raises a natural question: can one exploit both advantages within a single inference procedure?

The key idea behind VaSE is that the static and dynamic formulations are not only equivalent representations of the same Gaussian process, but can also be exchanged during inference. VaSE exploits this property by performing regression using the static GP formulation and subsequently converting the resulting posterior into an SPDE state representation for forecasting and posterior sampling. In this way, VaSE combines the strengths of kernel-based and SPDE-based inference.

\begin{figure}
    \centering
    \includegraphics[width=\linewidth]{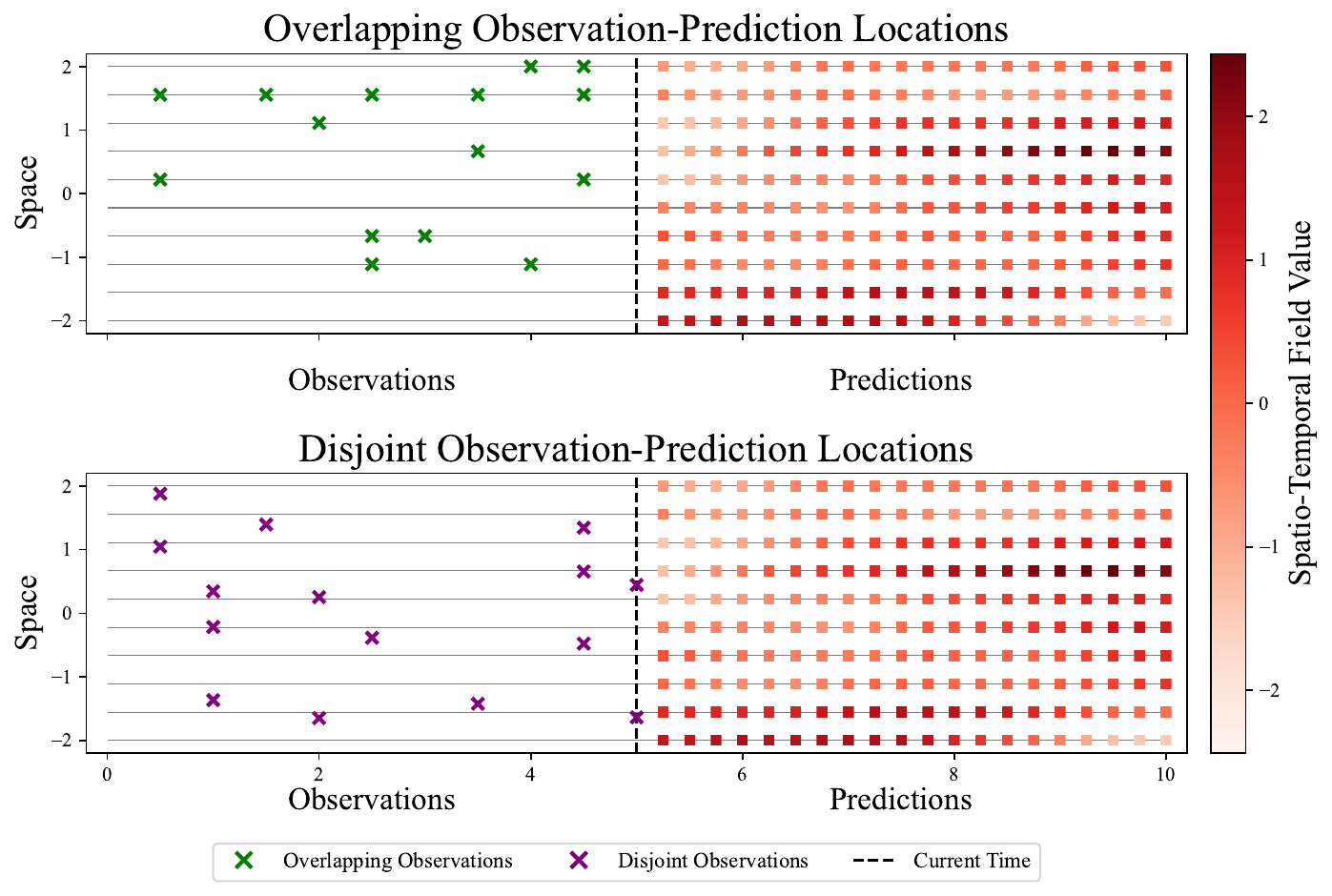}
    \caption{Plots illustrating overlapping (top) and disjoint (bottom) observation–prediction locations in one-dimensional space over time. Crosses denote observations, squares denote the prediction grid, the horizontal lines indicate the prediction locations, and the dashed vertical line marks the forecast boundary.}
    \label{fig:vase-posterior-motivation-1dspace}
\end{figure}

The benefits of this strategy are most apparent when the observation and prediction locations are largely disjoint, as illustrated in Figure \ref{fig:vase-posterior-motivation-1dspace}. In such settings, kernel-based GP regression can efficiently incorporate the observations, while direct SPDE inference becomes expensive because the state-space representation must simultaneously include both observation and prediction locations. VaSE does not introduce a new GP model; rather, it is a hybrid inference strategy that uses a static GP posterior as the initial condition for SPDE-based propagation. 

We illustrate the method using a spatio-temporal GP with a Mat\'ern-3/2 temporal kernel $k_{\text{Mat-}3/2}$ and a separable spatial kernel $k_s$,
\[
k_\text{full}((s,t),(s',t'))
=
k_s(s,s')k_{\text{Mat-}3/2}(t,t').
\]
Although we focus on this setting for simplicity of exposition, the construction applies generally to the class of separable kernels considered in Section~\ref{sec:dynamic-gp-spatio-temporal}.

The key observation underlying VaSE is that forecasting under the SPDE formulation requires only the posterior distribution of the SPDE state at the forecast boundary. Let $T$ denote the final observation time and $S^*$ the prediction locations. Rather than performing sequential inference over the entire spatio-temporal prediction grid, it suffices to obtain the posterior distribution of the SPDE state at $(S^*,T)$ and subsequently propagate this state forward using the SPDE dynamics.

To make such a conversion possible, the posterior distribution must be expressed in terms of the SPDE state variables. For a Mat\'ern-3/2 temporal kernel, the associated SPDE state consists of both $f$ and its temporal derivative $\partial_t f$. More generally, for a Mat\'ern-$(p+1/2)$ kernel, the state space consists of $f$ and $p$ partial derivatives of $f$ with respect to time. Since Gaussian processes are closed under linear operators \citep{sarkka2011linear}, the augmented process $\boldsymbol{f} = [f, \partial_t f]^T$ remains Gaussian with covariance obtained by applying the corresponding differential operators to the kernel:
\[
\begin{bmatrix}
    f(s,t) \\
    \partial_t f(s,t)
\end{bmatrix}
\sim
\mathrm{GP}
\left(
\begin{bmatrix}
0\\
0
\end{bmatrix},
\begin{bmatrix}
k_\text{full}
&
\partial_{t'}k_\text{full}
\\
\partial_t k_\text{full}
&
\partial_t\partial_{t'}k_\text{full}
\end{bmatrix}
\right).
\]
Therefore, standard GP regression can be performed using this augmented kernel, yielding the SPDE state required to initialise future propagation.

Suppose that observations are available up to time $T$ and predictions are required for future times $t>T$ at locations $S^*$. After performing GP regression with the augmented kernel, we compute the posterior distribution
\[
[f(S^*,T),\partial_t f(S^*,T)]^T
\mid
\mathcal D
\sim
N(\boldsymbol m_{T|T},\boldsymbol P_{T|T}).
\]
This distribution constitutes the filtering distribution of the SPDE state at the forecast boundary. Once $\boldsymbol m_{T|T},\boldsymbol P_{T|T}$ has been obtained, the kernel representation is no longer required. Future predictions and posterior samples can then be generated efficiently using the SPDE propagation formulas. The resulting procedure is summarised in Algorithm \ref{alg:vase-spacial-temp-disjoint} below. 

\begin{algorithm}[ht]
\caption{VaSE for spatio-temporal inference with disjoint observation locations.}\label{alg:vase-spacial-temp-disjoint}
\begin{algorithmic}[1]
\Require Spatio-temporal GP $\boldsymbol{f}$ with separable kernel $\boldsymbol{k}$. Current time $T$. Test locations $S^*$. Observations $\mathcal{D}$.
\State Augment the GP to $\boldsymbol{f} = [f, \partial_t f, \ldots, \partial_t^m f]^T$ with kernel $\boldsymbol{k}$ defined in Section \ref{sec:dynamic-gp-spatio-temporal}. 
\State Perform GP regression using the augmented kernel and compute the
posterior distribution of the SPDE state
$\boldsymbol f(S^*,T)\mid\mathcal D$.
\State Extract the posterior mean and covariance, denoted $\boldsymbol{m}_{T|T}$ and $\boldsymbol{P}_{T|T}$.
\State Propagate forward to time $t^*$ using the SPDE dynamics:
\[
\boldsymbol{m}_{t^*|T} = \Phi_{\text{full}}\, \boldsymbol{m}_{T|T}, \qquad
\boldsymbol{P}_{t^*|T} = \Phi_{\text{full}}\, \boldsymbol{P}_{T|T}\, \Phi_{\text{full}}^T + Q_{\text{full}}.
\]
\State Draw a posterior sample at $(S^*, t^*)$ via
\[
f^{(i)}(S^*, t^*) = G \left( \boldsymbol{m}_{t^*|T} + \boldsymbol{P}_{t^*|T}^{1/2} \boldsymbol{\eta}^{(i)} \right), \qquad
\boldsymbol{\eta}^{(i)} \sim N(0, I),
\]
where $G = I_{\text{space}} \otimes [1,\,0]$ extracts the physical process from the augmented state.
\end{algorithmic}
\end{algorithm}

Under Algorithm \ref{alg:vase-spacial-temp-disjoint}, drawing posterior samples then amounts to propagating the posterior moments $(\boldsymbol m_{T|T},\boldsymbol P_{T|T})$ and sampled realisations of the SPDE state forward using the Kalman prediction recursions. The key computational advantage of VaSE, therefore, is that regression is performed only at the forecast boundary $(S^*,T)$, while future predictions are generated through SPDE propagation. Consequently, the regression cost is reduced to $O(N^3 + N^2M_s + NM_s^2)$, whereas each posterior sample can be propagated forward at cost $O(M_s^2M_t)$. The resulting complexity decomposition is summarised in Table \ref{tab:comp_cost_vase}.

\renewcommand{\arraystretch}{1.2}
\begin{table}[ht] 
\centering

\begin{tabular}{|c|c|c|c|}
\hline
\textbf{Method} & \textbf{Regress } & $J$ \textbf{Samples } & \textbf{Total } $O(\cdot)$ \\
\hline

\text{Vanilla}
& $N^3 + N^2 M_s M_t + N M_s^2 M_t^2$
& {$JM_s^3 M_t^3$}
& {$N^3  + N^2 M_s M_t + N M_s^2 M_t^2+ JM_s^3 M_t^3$} \\
\hline

\text{SPDE}
& {$U_s^3 N_t$}
& {$JM_s^2 M_t$}
& $U_s^3 N_t + JM_s^2 M_t$ \\
\hline

\text{VaSE}
& {$N^3 + N^2 M_s + N M_s^2 $}
& {$JM_s^2 M_t$}
& {$N^3+ N^2 M_s + N M_s^2  + JM_s^2 M_t$} \\
\hline

\end{tabular} 
\caption{Computational cost comparison with regression and sampling phases separated.}
\label{tab:comp_cost_vase}
\end{table} 

Table \ref{tab:comp_cost_vase} highlights the complementary strengths of the three approaches following notations in Section \ref{sec:dynamic-gp-complexity}. Vanilla GP inference incurs substantial costs in both regression and posterior sampling on large spatio-temporal prediction grids. SPDE-GPs retain efficient sampling complexity but may suffer from expensive regression when the state must simultaneously include a large number of observation and prediction locations. VaSE combines the favourable components of both methods by retaining the static GP regression cost while preserving the linear-in-time sampling complexity of SPDE propagation.

In particular, when the observation and prediction locations are largely disjoint, the quantity $U_s = N_s+M_s$ may substantially exceed $N$. In this regime, the SPDE regression cost can dominate, whereas VaSE avoids this burden by performing regression entirely within the static formulation before switching to the SPDE representation for forecasting and posterior sampling.

\section{Experiments} \label{sec:experiments}

We validate our analysis with an empirical investigation. We evaluate the three approaches on a posterior sampling task for a spatio-temporal Gaussian process $f(s, t)$ with a separable kernel
\[
k((s, t), (s', t')) = k_s(s, s')\, k_t(t, t'),
\]
where $k_s$ is a two-dimensional RBF kernel and $k_t$ is a Mat\'ern-$3/2$ kernel admitting the dynamic formulation described in Section \ref{sec:dynamic-gp-spatio-temporal}. A synthetic ground truth is generated on a spatial grid of size $M_s = 15^2$ over a temporal interval $t \in [0,10]$ with increment $0. 05$ using the SPDE formulation. Observations are collected over $t \in (0,5]$ at regularly spaced times of increment $0.5$, with additive Gaussian noise. The regular observation time increment is only used, without the loss of generality, to simplify the SPDE method's implementation. We consider two observation scenarios:
\begin{itemize}
    \item \textbf{Overlapping Observation}: random locations from the prediction grid;
    \item \textbf{Disjoint Observations}: random locations inside the considered domain.
\end{itemize}
In both cases, the total number of observations $N$ ranges from $400$ to $2000$, with observations evenly distributed across observation times. The task is to draw $J=10$ posterior samples over $t \in [5,10]$ at increment $0. 05$ on the full spatial grid.

\begin{figure}
    \centering
    \includegraphics[width=\linewidth]{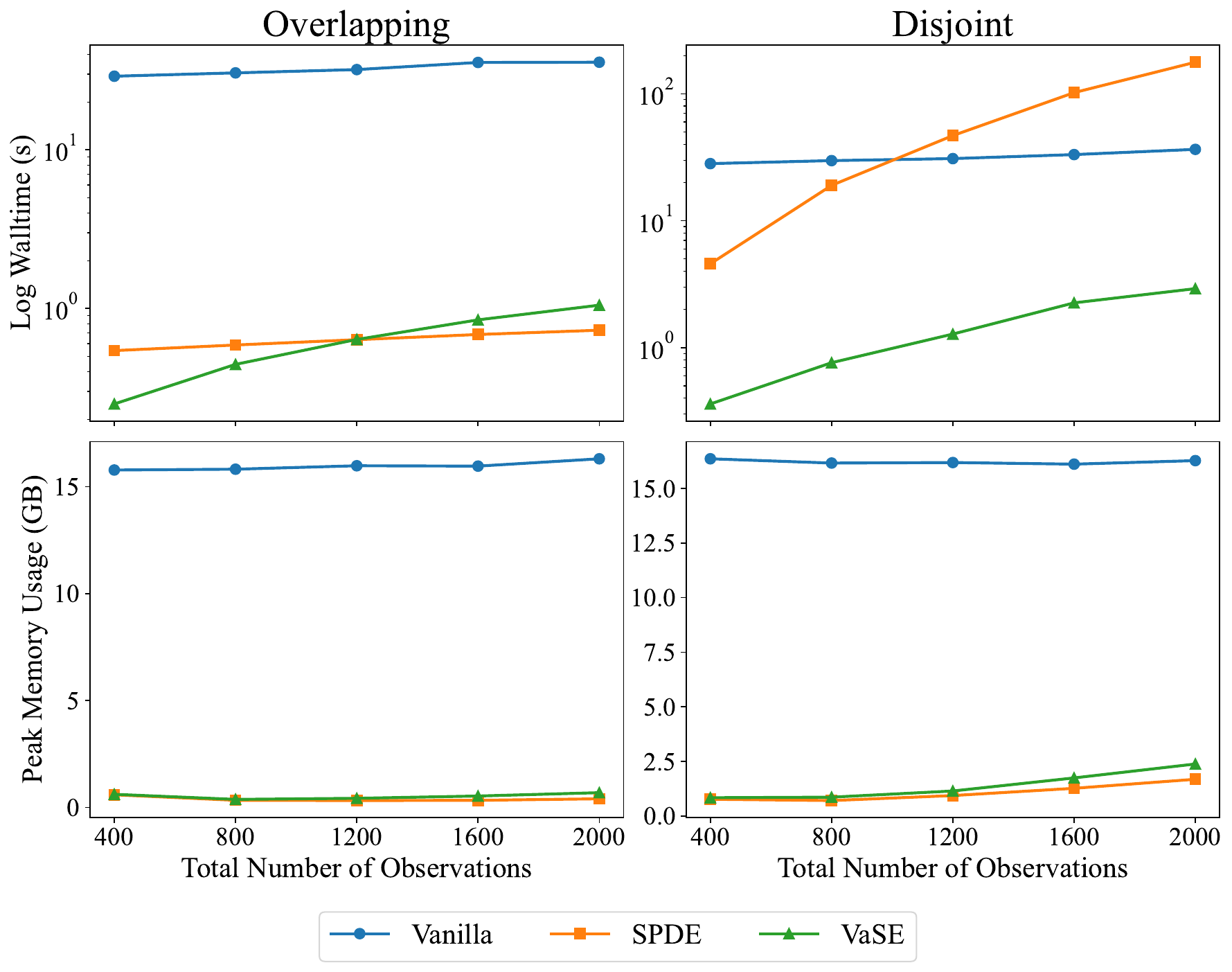}
    \caption{Median (out of three iterations) runtime on a log scale (top) and peak memory usage (bottom) for posterior sampling vs. number of observations with vanilla, SPDE, and VaSE methods. Left: overlapping observation-prediction locations; disjoint: disjoint observation-prediction locations.}
    \label{fig:sampling_benchmark}
\end{figure}

As shown in the results of Figure \ref{fig:sampling_benchmark}, VaSE is consistently better than the vanilla method, and is noticeably better than the SPDE method in the disjoint observation-prediction location setting. In the disjoint setting, we also notice that the SPDE method's walltime overtakes the vanilla method as the number of observations increases. These computational cost scaling aligns with the results in Table \ref{tab:comp_cost_vase}. Furthermore, both SPDE and VaSE have similar peak memory usage, while the vanilla method has a much higher usage due to the need to operate on the Gram matrix of all prediction locations. 

\section{Conclusions}

We presented the Vanilla-SPDE Exchange (VaSE) framework, which offers exact Gaussian process inference with the ability to leverage both the static and dynamic perspectives at once, and provide one case study of spatio-temporal posterior sampling under disjoint observation-prediction locations where such a hybrid inference is desirable. Our computational costs analysis and empirical evidence both demonstrate the benefits of VaSE over existing methods.

\begin{acknowledgement}
RZ is supported by EPSRC-funded STOR-i Center for Doctoral Training (grant no. EP/S022252/1). RZ, LA and EC are supported by the ARC ITRH for Transforming energy Infrastructure through Digital Engineering (TIDE), Grant No. IH200100009. 
\end{acknowledgement}

\bibliographystyle{spmpsci}
\bibliography{main}

\end{document}